# On the Accuracy of Edge Detectors
# in Number Plate Extraction

Bashir Olaniyi SADIQ[1], Emmanuel Okechukwu OCHIA[2],
Olayinka Sikiru ZAKARIYYA[3], Abdulazeez Femi SALAMI[3]

[1]Department of Computer Engineering, Ahmadu Bello University, Nigeria
[2]Department of Electrical and Computer Engineering, University of Calgary Canada
[3]Department of Electrical Engineering, University of Ilorin, Nigeria

bosadiq@abu.edu.ng, okechukwu.ochia1@ucalgary.ca,
zakariyya.os@unilorin.edu.ng, salami.af@unilorin.edu.ng

**Abstract.** Edge detection as a pre-processing stage is a fundamental and important aspect of the number plate extraction system. This is due to the fact that the identification of a particular vehicle is achievable using the number plate because each number plate is unique to a vehicle. As such, the characters of a number plate system that differ in lines and shapes can be extracted using the principle of edge detection. This paper presents a method of number plate extraction using edge detection technique. Edges in number plates are identified with changes in the intensity of pixel values. Therefore, these edges are identified using a single based pixel or collection of pixel-based approach. The efficiency of these approaches of edge detection algorithms in number plate extraction in both noisy and clean environment are experimented. Experimental results are achieved in MATLAB 2017b using the Pratt Figure of Merit (PFOM) as a performance metric.

**Keywords:** Edge Detection, Number Plate Extraction, Pixels.

## 1. Introduction

Edge detection is a basic but fundamental step in image prepossessing and analysis. Improper technique in the extraction of edges can lead to the identification of false and broken edges amongst others (Sadiq et al, 2015). In literature, some of the most common edge detection techniques are the Sobel, Prewitt, Laplacian, Canny, Robert edge detectors. These existing edge detection algorithms have a generalized concept of using a single value pixel processing technique by calculating a value that shows the edge magnitude and orientation. Amongst these existing edge detectors, the canny edge detection algorithm has been proven to function better in the detection of edges than the Sobel, Prewitt, Robert, and Laplacian. This is because researchers such as Almadhoun (2013), Bhardwaj (2013), Sadiq et al. (2016) have shown that the canny edge detection algorithm minimizes broken, false and thick edges whilst reducing the presence of noise of certain degree. However, the computational time is not reduced using the Canny edge detector. This is why the authors in the work of Agrawal and Chawla (2017) prefer the use of the Sobel edge detector in the extraction of the license plate. But this edge detector failed because of some factors that affect the number plate such as lighting conditions, movement of vehicles and climatic changes. The use of the Laplacian edge detector has also been employed for character extraction in the license plate as presented by the authors in the work of Utkarsh et al. (2017). Their technique used Otsu method for noise removal in the captured image. The captured image is then binarized and the license plate is located using the Laplacian operator. However, using the Laplacian operator, the location accuracy is lower due to the size of the filtering kernel. Therefore, a collection of pixel approach is required to explore the larger area to overcome noise and consider the global structure of edges with a view to reducing false and broken edges in a reasonable time (Sadiq et al., 2015). Nonetheless, the technique presented by the authors in (Sadiq et al., 2015) works best for clean images only. In order for their technique to function properly, the authors in (Sadiq et al., 2016) augmented (Sadiq et al., 2015) with a Particle Swarm Optimization (PSO) based median filter with a view to removing the noise present in the image. These aforementioned techniques of edge detections are based on single value pixel with the exception of the works of Sadiq et al. (2015) and Sadiq et al. (2016) that used a collection of pixel-based approach. Also, the detectors are applied to either the colored images as they are originally captured or converted to grey scale before detection of edges. The formula used to convert images from colored to grey scale is presented in equation (1)

$$Grey(i,j) = 0.2989 * R + 0.587 * G + 0.114 * B \qquad (1)$$

where Grey(i,j) is the converted grey scaled image and R, G, B are the Red-channel, Green-channel, and Blue-channel of the colored image (Junaid et al., 2017).

The Eq. (1) finds application when using the traditional edge detectors like the Sobel, Prewitt and Robert (Jahanzeb and Siddiqui, 2013; Padma et al, 2017). The images are first converted to grey scale before performing the detection of edges. However, during the course of this experimental work, it was observed that the Canny edge detection algorithm outperformed the Sobel, Prewitt and Robert in terms of accuracy of edges detected. Therefore, the best candidate amongst these four-edge detection algorithms which is the canny that uses a single value pixel approach was selected with a view to comparing it with that used the collection of pixel-based approach.

License plate extraction is achieved using the edge statistics and morphology as presented by the authors in the works of Jun-We et al. (2002); Bai and Liu (2004); Samiul and Monirul (2016), Sarbjit (2016), Padma et al. (2017), Chaetan et al. ( 2017), Ohnmar et al. (2017), Guilherme et al. (2017), Faizal et al. (2018), Kumar et al. (2018). Their techniques function best on grey scale images. The use of morphology to characterize a pixel as an edge requires convolving a spatial filter with the image in both the vertical and horizontal axis. This is termed as vertical and horizontal edge detectors. Morphology actually means the act of structuring. This concept usually involves three key processes which are:

    a.   Smoothening Operation (SO)
    b.   Dilation Operation (DO)
    c.   Erosion Operation (EO)

The underlying equations that characterize these processes are defined as:

$$SO_{a,b}(I(x,y)) = \frac{1}{ab} \sum_{i=-b/2}^{b/2} \sum_{j=-a/2}^{a/2} I(x+i, y+j) O_{a,b}(i,j) \qquad (2)$$

$$DO_{a,b}(I(x,y)) = \max_{|i| \leq a/2, |j| \leq b/2} I(x-i, y-j) O_{a,b}(i,j) \qquad (3)$$

$$EO_{a,b}(I(x,y)) = \min_{|i| \leq a/2, |j| \leq b/2} I(x-i, y-j) O_{a,b}(i,j) \qquad (4)$$

These operations are performed on an input image denoted by I(x,y). The structuring element is denoted as $O_{a,b}$ with the size of a $x$ b, where a and b are odd numbers and larger than zero. However, the fundamental drawback of the morphological edge detectors is that the structuring element can only serve the same direction feature. Once the features are of different directions, the morphological edge operators cannot function effectively. Thus, making it not suitable as a standalone technique in identifying edges in number plate extraction.

The steps involved in the detection of edges are itemized as (Gonzalez, 2007; Bharat et al., 2012; Allam, 2012; Sarbjit and Sukhvir, 2014; Khalid and Adarsh, 2014; Chetan and Ajit, 2014; Samarth, 2015; Reji and Dharun, 2015; Soumyadip et al., 2017; Isra and Gokulanathan, 2017; Ameya and Shubhangi, 2017):

   a.   Smoothening: noise removal in the captured image is essential before detection of edges. This should be done with care in order not to destroy the true edge. Some of the types of noise that occurs in the captured images are but not limited to salt & pepper, Gaussian and motion blur.
   b.   Enhancement: Most often but not always, filters are applied to the images to enhance the quality of the edges in the images otherwise known as sharpening.
   c.   Detection: This involves the processing of using a thresholding scheme to select with pixels will be considered as edges and which will be discarded as noise.
   d.   Localization: this deals with the process of edge thinning and linking.

The intensity changes in pixels that constitute an edge are presented in Fig 1 (Isra and Gokulanathan, 2017).

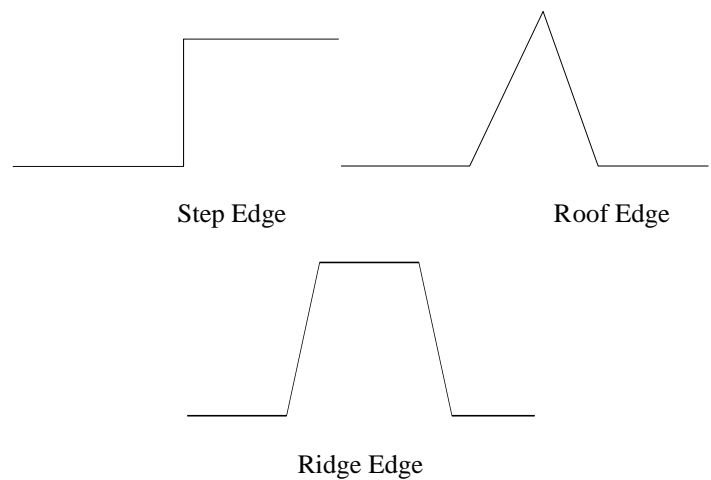

Step Edge                              Roof Edge

Ridge Edge

**Fig 1**. The Intensity Chane in Pixel

Based on the review of the existing edge detection method in the number plate extraction system, edge detection algorithms can be categorized into two forms; which are those based on a single value pixel approach and those based on a collection of pixel-based approach. As such, algorithms that used both approaches will be replicated and compared to ascertain the most effective as applied to number plate extraction.

The purpose of this research work is to determine the suitability of the edge detection algorithm using collection of pixel-based approach in number plate extraction. This is because based on existing literature, the single pixel-based edge detection approach could not function in noisy environment and are known to produce false and broken edges which are not good in number plate extraction.

The type of number plate actually affects the efficiency of an edge detection algorithm. These number plates can be categorized as either clean, dirty or faded as depicted in Fig 2.

The number plate extraction serves a vital role in the intelligent transport system because of the increasing number of vehicles on the roads performing different day to day operation. With the increase in the number of vehicles that ply the roads, there is also an increase in criminal activities. This makes the number plate extraction system important to the law enforcement agent (Junaid et al, 2017). Amongst other uses of the

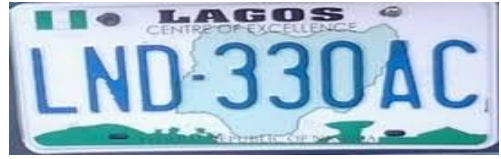

(a) Clean Number Plate

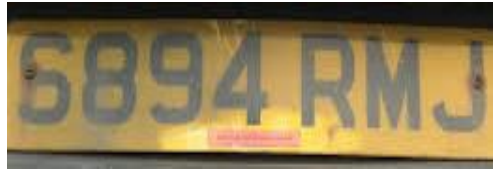

(b) Dirty Number Plate

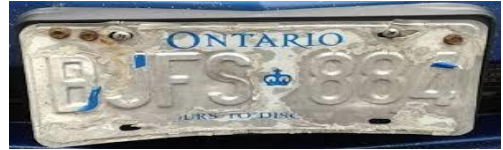

**(c)** Faded Number Plate

**Fig 2.** Types of Number Plate

license plate extraction is the parking facilities, toll gate ticketing, and traffic management. A quantitative way to determine if the edges in the number plate are identified properly is with the use of the Pratt Figure of Merit (PFOM) (Gonzalez et al, 2016). The equation of the PFOM is

$$h = \frac{1}{Max(k_I, k_A)} \sum_{i=1}^{N_A} \frac{1}{1 + nl^2(p)} \qquad (4)$$

where:   $k_I$ is the number of actual edges, $k_A$ is the number of detected edges, n is a scaling constant set to 1/9, l(k) denotes the distance from the actual edge to the corresponding detected edge.

The Pratt Figure of Merit measures values between 0 and 1 as the upper and lower limit respectively. The closer the value is to 1 simply implies the detection of more accurate edges.

The flow chart of the generalized number plate extraction system identifying the edge detection as a pre-processing stage in bold is depicted in Fig 3.

The remaining facet of the paper are ordered as follows: section two presents the materials and methods used, steps to pre-process the image based on the flow chart presented in Figure 2 as well as the data used in the prepossessing stage. Section three shows the result of works and discussion drawn from their method used. Section four conclude on the accuracy of the prepossessing stage in the number plate extraction.

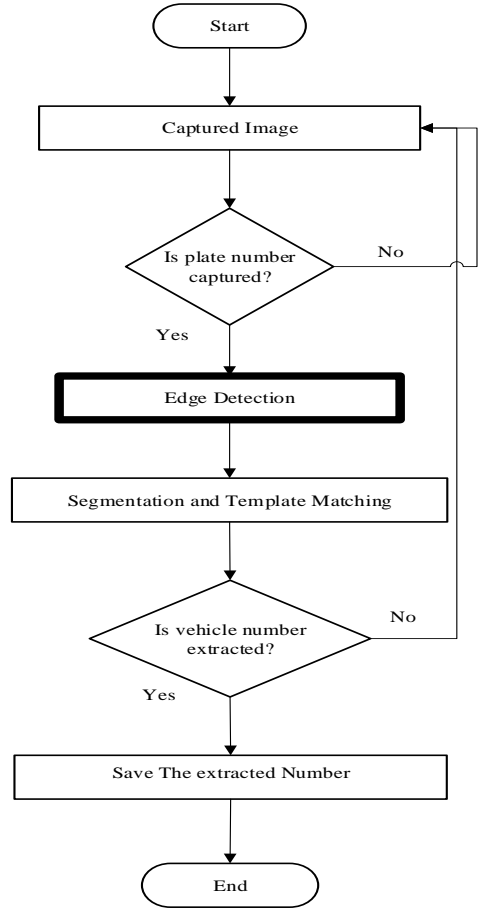

**Fig 3**. Flow Chart of the Generalized Number Plate Extraction System

## 2. Generalized Methodology

Based on the surveyed literature, edge detectors can be generalized in two forms:
1. Those that use single value pixel approach to categorize an edge.
2. Those that use a collection of pixel-based approach to categorizing an edge.

These approaches are replicated using the underlisted steps and are applied to the number plate in both clean and noisy environment. It is important to note that the Canny, Sobel, Roberts, Laplacian and Roberts are categorized as single pixel-based approach while that presented in the work of Sadiq et al. (2015) and Sadiq et al. (2016) is categorized as a collection of pixel-based approach.

The materials used in this work are:
**a**. A computer system with 4GB RAM, 2.3GHz processor, and MATLAB 2017b installed software used to run the edge detection algorithm.
**b**. A 5 Mega Pixel digital camera placed at a distance of 10m used to capture the image (clean number plate).

Noise intensity of 50% was added to the captured images in MATLAB 2017b with a view to depicting the worst-case scenario of the noise level in a captured data. This is thus presented.

*filename = uigetfile('\*.tif,\*.jpg');*
*data=im2double(imread(filename));*
*i=input('please input the percentage of Noise=');*
*noisy= imnoise(data,'speckle',i);*

The addition of 50% noise to the sample of the image data collected is depicted in Fig 4.

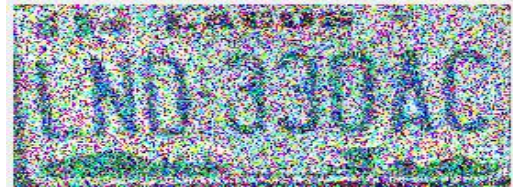
(a) Clean Number Plate with Noise

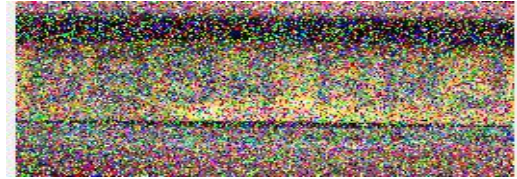
(b) Dirty Number Plate with Noise

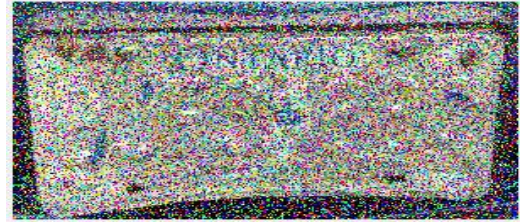
(c) Faded Plate Number with Noise

**Fig 4.** Number plate with Noise

The addition of noise to the image data shows that the captured images are not noise free and the noise varies in degree depending on the distance of capture, the motion of the vehicle, lightening condition amongst others. The type of noise added to the data collected is the impulse noise. This type of noise appears in form of black and white dots in an image. The noise occurs in the image due to sharp and sudden change in image signal which has been categorized as a step, ramp, roof and ridge edges (Jahanzeb and Siddiqui, 2013; Padma et al., 2017). The edge profile modeling based on the different edge response has been presented by the authors in Sadiq et al. (2015).

The detailed steps of detecting the edges in the number plate extraction based on the two broad classes are as follows:

a. Capture the number plate using a digital camera.
b. Replication of algorithms that used the single pixel value approach
c. Replication of the work presented in Sadiq et al, (2015) and Sadiq et al, (2016) because these are one of the few works that used a collection of pixel-based approach.
d. Reconstruct the image from a noisy environment to a clean environment using an averaging filter based on Eq. (5) and (6).

$$\text{Avf}(x, y) = \sum_{j=-1}^{1} \sum_{i=-1}^{1} 1 \times g(x+i, y+j) \tag{5}$$

$$\text{Avf}(x, y)\_\text{normal} = \frac{1}{\sum_{j=-1}^{1} \sum_{i=-1}^{1} 1} \sum_{j=-1}^{1} \sum_{i=-1}^{1} 1 \times g(x+i, y+j) \tag{6}$$

where Avf(x,y) is the filtered image after addition of noise, Avg(x,y) is the normalized final output image of the filter while g(x,y) is the noisy image.

The need for normalizing is very important in order to keep the image pixel values between the value of 0 and 255.

Enhance the brightness of the image with a view to detecting sharp edges.

    1. Given the smoothed image as input image g(i,j)

A logarithmic transformation is one of the easiest ways to increase the brightness on an image. This is controlled by a parameter denoted as presented in [31]. The equation of the enhancement filter used is:

$$E(i, j) = m * g(i, j)^{\sigma} \tag{7}$$

where E(x,y) is the enhanced image, g(i,j) is the input image.

The value of m is a constant parameter and, in this case, set as 1. $\sigma$ is a control parameter. It defines the brightness enhancement level. In this paper, the value of $\sigma$ is intuitively set to 0.2 which means the brightness of the enhanced image is 20% that of the original image.

    2. [a.b] = size (image)
    3. double_image = im2double(image)
    4. for i =1:a
          for j = 1:b
              E(i,j) = 1*power(g(i,j).^0.2
          end
        end

  e.  Detect the edge in the reconstructed clean image using the steps presented.

  f.  Using the Pratt figure of Merit (PFOM) Equation, determine the connected edges in the extracted edge. The equation of the PFOM is presented in Eq. (4)

  g.  Proceed to segmentation and template matching

## 3. Experiment

Sequel to corrupting the captured images with 50% impulse noise, edges of the number plates were extracted to ascertain the efficiency of both the single and collection of pixel-based approach. It was noticed that both the existing edge detection algorithms such as the Robert, Canny, Prewitt, Laplacian & Sobel that used the single pixel-based approach did not yield any meaningful result as depicted in Fig 5. So also, was those based on the collection of the pixel-based approach in Sadiq et al, (2015) and Sadiq et al, (2016) despite the use of an optimization filter as presented in their work. This was due to the fact that this algorithm cannot function with a high level of noise.

The result presented in Figure 6 was the output result of the Canny edge detector that was reported in the literature to be the best amongst the existing once (Allam, 2012; Samarth, 2015) for single collection of pixel-based approach. This implies that all these algorithms are not suitable for the very high level of noise. As such, the noise in the captured images was decreased in a range of 2% until a noise value of 40%. Even at a noise level of 40%, it was noticed that only edges in the clean number plate were detected exactly. A further decrease in the noise level of images was considered to determine exactly at what percentage noise level will these algorithms fail completely. It was noticed that detection of edges was feasible at the intensity of noise 30% and below. Figure 6 presents the test result of the edge detection algorithms.

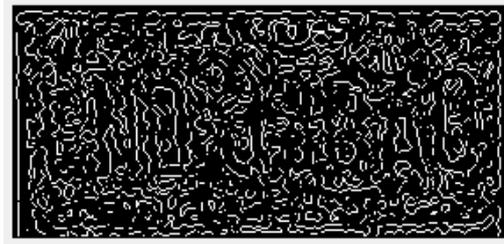

Detected Edges of Clean Number plate at 50% noise

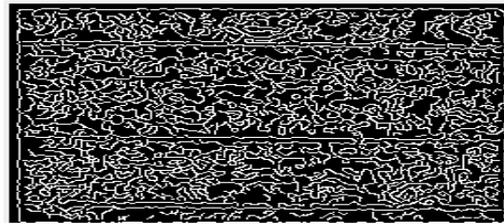

Detected Edges of Dirty Number Plate at 50% noise

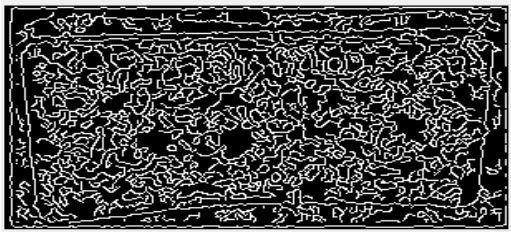

Detected Edges of Faded Number Plate at 50% noise

**Fig 5.** Detection of Edges in Number Plate with High-Level Noise using Collection of Pixel Based
Approach

From the test result, it shows that the technique of detecting edges as presented in
Sadiq et al, (2015) is most suitable to detect edges in number plate extraction in
comparison to the Sobel, Canny, Prewitt, Robert, and Laplacian. However, this is
achievable only if the noise level is below 30%.

Therefore, in order to efficiently detect edges in images of both noisy and clean
images, there is need employ techniques such as deep learning amongst others to the
argument the traditional edge detectors in number plate extraction.

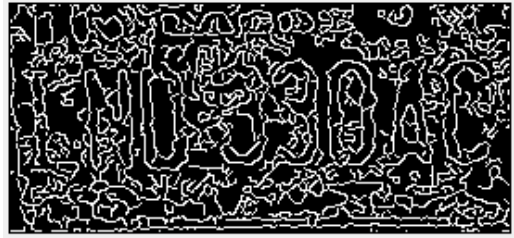

Canny Edge Detector at 40% Noise

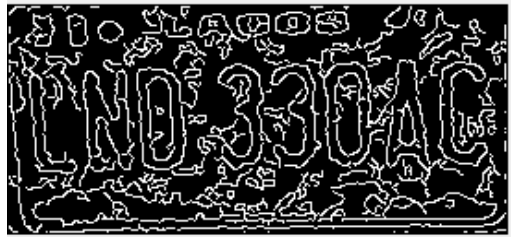

Canny Edge Detector at 30% Noise

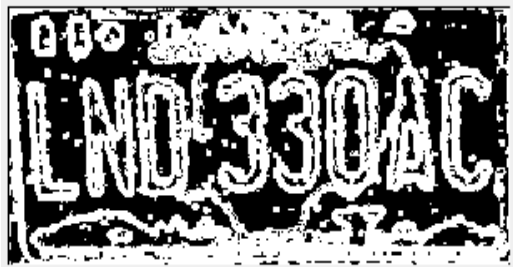

Sadiq et al. (2015) at 30% Noise

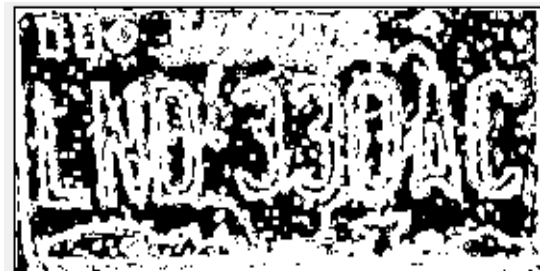

Sadiq et al. (2015) at 40% Noise

**Fig 6.** Test Result of the categories of Edge Detection Algorithms Under Noise for both Single and Collection of pixel-based approaches

Pratt figure of Merit states that the closer the value obtained to 1, the better the edges are detected. Therefore, the collection of pixel-based approach presented in the work of Sadiq et al. (2016) is closer to 1. However, only speckle noise was considered in the development of the algorithm and it is limited to a noise level of 30%. Speckle noise was only considered because it mostly occurs in the images as a result sharp and sudden changes in image signal due to dust particles during the image capturing stage or corrupted transmission channel. This is inevitable because the cameras to capture plate numbers are usually placed outdoors.

**Table 1**. Pratt Figure of Merit (PFOM) Table.

| Edge Detection | Type | Image without noise | Image with noise (30%) |
|---|---|---|---|
| Prewitt Edge Detection Algorithm | Single Value Pixel-based approach | 0.4091 | 0.3891 |
| Roberts Edge Detection Algorithm | Single Value Pixel based approach | 0.4181 | 0.2401 |
| Canny Edge Detection Algorithm | Single Value Pixel-based approach | 0.8572 | 0.4608 |
| Laplacian Edge Detection Algorithm | Single Value Pixel based approach | 0.6908 | 0.2811 |
| Sobel Edge Detection Algorithms | Single Value Pixel-based approach | 0.4101 | 0.4011 |
| Sadiq et al, (2015) Edge Detection Algorithms | Collection of pixel-based approach | 0.8480 | 0.8321 |
| Sadiq et al, (2016) Edge Detection Algorithms | Collection of pixel-based approach | 0.8546 | 0.8546 |

## 4. Open Issues

Some of the problems identified with the existing works in edge detection algorithms in number plate extraction are:
1. Due to the cruciality and importance of the number plate extraction, enhanced noise filtering techniques are needed to remove noise. This will improve the performance of the edge detection algorithm presented using both single value pixel approach and collection of pixel value approach.
2. Most of the edge detection algorithms work best with the speckle noise. However, motion deblurring filtering technique should be considered because images are captured while the cars are in motion. As such, there is a need to develop an algorithm that will be robust to more than one type of noise and of varying degree.
3. The font size of the car number plate affects the performance of the edge detection algorithms. Improved edge detection algorithms should be developed in order to adapt to the varying size of different number plates. As of now, there is no multiple number plate extractor that functions based on different number plate designs with respect to different countries. For the purpose of extraction, each number plate extractor is designed based on the requirement of the applied area.
4. Enhancement filters are highly required in the aspect of faded and dirty plate numbers. This limits the performance of the edge detection algorithms. Optimization techniques can be used together with filters to improve the localization accuracy of edges.

## 5. Conclusion

This paper presented the edge detection techniques as a pre-processing stage and fundamental of the number plate extraction system. This is due to the fact that the identification of a particular vehicle is achievable using the number plate because each number plate is unique to a vehicle. As such, the characters of a number plate system that differs in lines and shape can be extracted using the principle of edge detection. The paper compared the different types of edge detection algorithm for number plate extraction. Based on the test carried out, open issues and areas of further improvement were highlighted. Amongst the edge detectors used, the one that used a collection of pixel-based approach performed better than others using visual comparison. Also, quantitative performance was obtained using the PFOM that shows that a collection of pixel-based approach obtaining a PFOM of 0.8480 and 0.8546 in the work of Sadiq et al. (2015) and Sadiq et al. (2016) respectively performed better than the single pixel-based approach that obtained a PFOM of 0.8321 and 0.8546 in the work of Sadiq et al. (2015) and Sadiq et al. (2016) respectively. Further work can concentrate on the open issues highlighted.